\documentclass[conference]{IEEEtran}
\usepackage{cite}
\usepackage{amsmath,amssymb,amsfonts}
\usepackage{algorithmic}
\usepackage{graphicx}
\usepackage{textcomp}
\usepackage{xcolor}
\usepackage{qtree}
\def\BibTeX{{\rm B\kern-.05em{\sc i\kern-.025em b}\kern-.08em
    T\kern-.1667em\lower.7ex\hbox{E}\kern-.125emX}}
\begin{document}

\title{Target Identification and Bayesian Model Averaging with Probabilistic Hierarchical Factor Probabilities
}

\author{\IEEEauthorblockN{Bill Basener}
\IEEEauthorblockA{\textit{Professor of Data Science} \\
\textit{University of Virginia, School of Data Science}\\
Charlottesville, VA \\
wb8by@virginia.edu}
}

\maketitle

\begin{abstract}
Target detection in hyperspectral imagery is the process of locating pixels from an image which are likely to contain target, typically done by comparing one or more spectra for the desired target material to each pixel in the image.  Target identification is the process of target detection incorporating an additional process to identify more specifically the material that is present in each pixel that scored high in detection.  Detection is generally a 2-class problem of target vs. background, and identification is a many class problem including target, background, and additional know materials.  The identification process we present is probabilistic and hierarchical which provides transparency to the process and produces trustworthy output.  This process of target identification is central to the NINJA near real time hyperspectral image processing software, which is currently being incorporated for automated processing of images on board a UAS using GPUs.

In this paper we show that target identification has a much lower false alarm rate than detection alone, and provide a detailed explanation of a robust identification method using probabilistic hierarchical classification that handles the vague categories of materials that depend on users which are different than the specific physical categories of chemical constituents.  It should be no surprise that target identification outperforms target detection because the identification process can include spectra for correlated materials that can cause false alarms in detection, directly eliminating false alarms if suitable spectra and methods are used.

Identification is often done by comparing mixtures of materials including the target spectra to mixtures of materials that do not include the target spectra, possibly with other steps. (band combinations, feature checking, background removal, etc.)  Standard linear regression does not handle these problems well because the number of regressors (identification spectra) is greater than the number of feature variables (bands), and there are multiple correlated spectra.  Our proposed method handles these challenges efficiently and provides additional important practical information in the form of hierarchical probabilities computed from Bayesian model averaging
\end{abstract}


\section{Introduction}
\label{introduction}
Target detection is the process of determining pixels in an image that contain evidence of a target spectrum.  Typically, this is done as a 2 class problem with a target class and a non-target class.  A common detection algorithm is the ACE (Adaptive Cosine Estimator) which involves whitening each image pixel and target spectrum and then computing the correlation between the whitened pixel and target (equivalently, angle between these whitened spectra) described in~\cite{basener2017geometry}.  Identification is a more complex process, usually done using a form of unmixing, or regression, to determine the mixture of material spectra present in a pixel.  This identification via unmixing is done by using linear regression to determine the best mixture of known material spectra (called a model) that approximates the observed pixel spectrum.

This target identification process was introduced in automated processing in~\cite{Basener2011, Basener489}, and has become an integral element of automated processing for example on board a UAS such as the GeoReplay-Lite and HyPert~\cite{AdlerGolden_osti_1638363_2020} software.  A major challenge in the identification by regression process is that there the libraries used can be extremely large, as the goal is to identify combination the material(s) present in the pixel out of thousands of known materials.  With many more regressor variables (spectra) than features (bands), this is an underdetermined regression problem.  One method proposed to deal with this is merging groups of similar spectra using a hierarchical clustering~\cite{Owechko2016, LoughlinManolakis2020}, and find the best unmxing model using spectral cluster means.  An alternative is to compute probabilities for many models which leads to probabilities for individual spectra or classes of spectra~\cite{Basener2017}, a process called model averaging which is the method we focus on in this paper.

Bayesian model averaging is an important method for averaging results for a variety of models with variation.  Used in multivariate linear regression, it provides a robust form of model selection which captures the uncertainty about model selection, providing a regularized regression model that is the average of the top models rather than picking a single best model along with a probability of inclusion for each input variable.  In this setting, each model is a least squares linear regression model and the models differ by the choice of subset of input variables, although of course different forms of variation in models could be used.

The fundamentals of Bayeisan model averaging are straightforward.  For a multivariate linear regression problem, let $M_i$ denote a model; that is a subset of the input variable used for linear regression.  Let $BIC_i$ be the Bayesian Information Criterion associated with this regression model.  Then the likelihood for this model is approximately
\begin{equation}
l(M_i) = e^{-BIC_i/2}
\end{equation}
From Bayes Theorem the posterior probability for this model is then
\begin{equation}
P(M_i)=\frac{l(M_i)Pr(M_i)}{\sum_k l(M_k)}.
\end{equation}
If our input variables are $X_1,...,X_n$ then the probability that $X_k$ should be included is equal to the sum of probabilities for all models that contain $X_k$,
\begin{equation}
P(X_k) = \sum_{M|X_k\in M} P(M).
\end{equation}
Note that this sum is over all models $M$ which contain $X_k$ as an input variable.  This is the Bayesian posterior probability for including $X_k$ in the model; the probability that the coefficient for $X_k$ is nonzero given the data.  The model averaged coefficient for $X_k$ is then
\begin{equation}
\beta^{\textrm{avg}}_k = \sum_{M|X_k\in M} P(M)\beta^{M}_k,
\end{equation}
where $\beta^{M}_k$ is the  coefficient of $X_k$ in model $M$.  These model averaged coefficients are weighted by model probabilities, and this process is a form of regularization creating the averaged model using these coefficients.

If the number of input variables is significant, it can (often is) computationally impossible or infeasible to compute these probabilities for every possible model.  The two main approaches to sampling are a Markov Chain Monte Carlo model composition ($MC^3$) approach using MCMC random sampling and an Occam's window approach where models are iteratively constructed and rejected when their likelihood is low relatively to the best previous model (details in~\cite{RafteryMadiganHoeting1997}).  Implementations are available in R~\cite{AminiParmeter2011}.

In this paper, we address the issue where multiple input variables represent the same underlying factor or cause.  If multiple variables are measuring the same underlying factor, their probabilities can compete giving results that are misleading or require manual interpretation which is not always possible.  It is often important to use multiple variables measuring the same underlying factor/cause, and the problem cannot generally be solved by choosing just one representative.  We give a first example in a classic context of crime data, and then focus on regression models for the important application of identifying materials our of mixtures in hyperspectral data.

We will refer to input variables that represent the same underlying factor\\cause\\information as \textit{overlapping variables}.  The definition is subjective, dependent on the goals of the regression analysis, an issue we will deal with particularly in the spectral data.

\section{Overlapping Variables in Crime Data}
A classic paper in the area is Bayesian Model Averaging for Linear Regression Models~\cite{RafteryMadiganHoeting1997}.  This paper used a classic (often criticized) dataset and model that historically influenced the question of whether crime is deviant behavior linked to the offender's psychological social, and family circumstances, or whether criminal activity is determined by rational choices based on costs and benefits of alternative opportunities.  This dataset includes 15 demographic factors from each state in the U.S. along with the crime rate for that state in 1960.

The posterior probability for inclusion from~\cite{RafteryMadiganHoeting1997} for each input variable in the Erlich crime data is shown in Table~\ref{ErlichPIPs}.
\begin{table}[htbp]
\caption{Posterior Probability of Inclusion for input predictor variables in crime data for 1960, expressed as a percentage. (UR = Unemployment Rate, PR = Participation Rate)}
\begin{center}
\begin{tabular}{|l|c|c|c|r|}
\hline
Predictor    & Occam's & $MC^3$ \\
             &  Window &         \\
\hline
\% of males age $14-24$     & 73   & 79    \\
South state indicator       & 2    & 17    \\
Mean years schooling        & 99   & 98    \\
Police expenditure, 1960    & 64   & 72    \\
Police expenditure, 1959    & 36   & 50    \\
Labor Force PR              & 0    & 6     \\
Males per 1,000 females     & 0    & 7     \\
State population            & 12   & 23    \\
nonwhites per 1,000         & 53   & 62    \\
UR urban males $14-24$      & 0    & 11    \\
UR urban males $35-39$      & 43   & 45    \\
Wealth                      & 1    & 30    \\
Income Inequality           & 100  & 100   \\
Prob. of Imprisonment       & 83   & 83    \\
Avg time served in prison   & 0    & 22    \\
\hline
\end{tabular}
\label{ErlichPIPs}
\end{center}
\end{table}

Our focus on situations where multiple input variables represent the same underlying factor or cause is illustrated in the police expenditure in 1959 and in 1960 variables.  Observe that the sum of probabilities for these two separate variables is one, and it seems these are two distinct variables capturing the same underly factor or information.  Raftery et al~\cite{RafteryMadiganHoeting1997} discuss the problem as follows.
\begin{quote}
The model averaging results for the predictors for police expenditures lead to an interesting interpretation.  Police expenditure was measured in two successive years, adn the measures are highly correlated ($r=.993$).  The data show clearly that the 1960 crime rate is associated with police expenditures, and that only one of the two measures ($X_4$ and $X_5$) is needed, but the do not say which measure should be used.  Each model in Occam's window contains one predictor or the other, but not both.  For both Occam's window and $MC^3$ $Pr[(\beta_4 \neq 0) \cup (\beta_5 \neq 0) | D]=1$, so the data provide very strong evidence with police expenditure.
\end{quote}

\section{Overlapping Variables in Hyperspectral Data}
\begin{figure}[ht]
\vskip 0.2in
\begin{center}
\centerline{\includegraphics[width=\columnwidth]{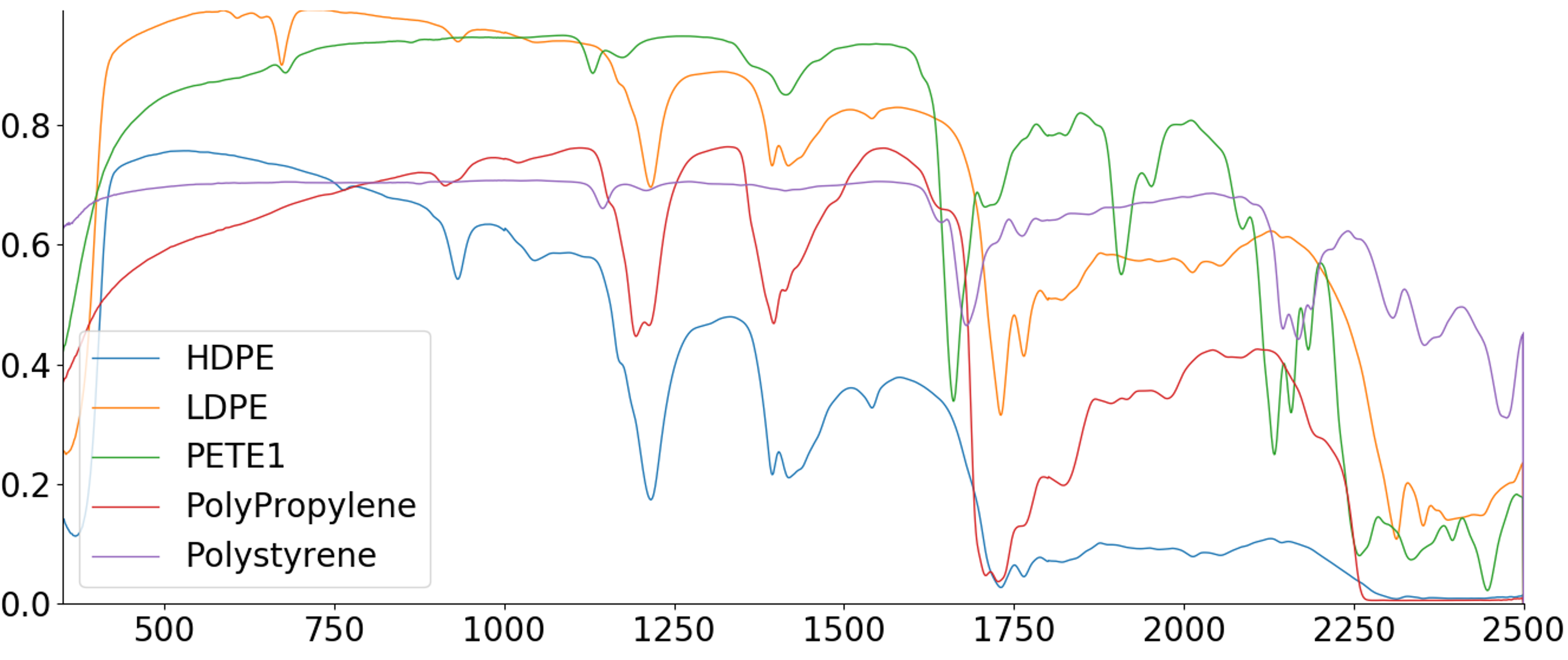}}
\caption{Spectra for five different polymer materials.}
\label{PolymerSpectra}
\end{center}
\vskip -0.2in
\end{figure}

A hyperspectral image is a digital image in which each pixel has not just the usual three (red, green, blue) colors, but many, usually more than one hundred, bands.  Each band is a measurement of light across a small range of wavelengths.  Some of our data will come from the NASA AVIRIS sensor whch has 480 bands ranging from 380nm to 2510nm, and each band collects light from a range of about 5nm.  The data can be `binned' combining neighboring bands to reduce noise and some bands are removed where the atmosphere (particularly water) blocks most or all of the light, and our AVIRIS data as a resulting 168 bands.  The data is converted to reflectance units, meaning that the value for each band in each pixel is the percent of light at the wavelengths for that band that were reflected off the ground at the location of the pixel.  The reflectance varies with wavelength due to interactions of the photons with molecular constituents and bonds in the material, and thus the reflectance spectrum provides information about the material present.  Spectra for five different polymer materials are shown in Figre~\ref{PolymerSpectra}.

Each pixel in a hyperspectral image collected from an aircraft is the measurement of light from a relatively large region on the ground - typically ranging from a 2m$\times$2m up to 10m$\times$10m depending on the altitude of the sensor.  Thus, there are usually multiple materials present in the ground region and the measured spectrum is a mixture of the spectra for those materials.  This phenomenology results in what is called linear mixing; if $x_1,...,x_n$ are the spectra (vectors whose length is the number of bands) for the materials present on the ground, the measured pixel spectrum will be
\begin{equation}
p=\sum_i a_i x_i + \textrm{err}
\end{equation}
where $a_i$ represents the \textit{abundance} of the material in the pixel; physically the fraction of light reflecting off that material out off all the light that reflected off the ground region for the pixel.  This is known as the linear mixing model.  There are factors that can create nonlinear mixing of spectra, for example translucent materials like plant leaves resulting in light passing through a leaf and reflecting off the ground, geometry of objects where light reflects off two or more materials, or intimate mixtures powder coatings), for example.  However, the linear mixture model is usually a good practical approximation.

\begin{figure}[ht]
\vskip 0.2in
\begin{center}
\centerline{\includegraphics[width=\columnwidth]{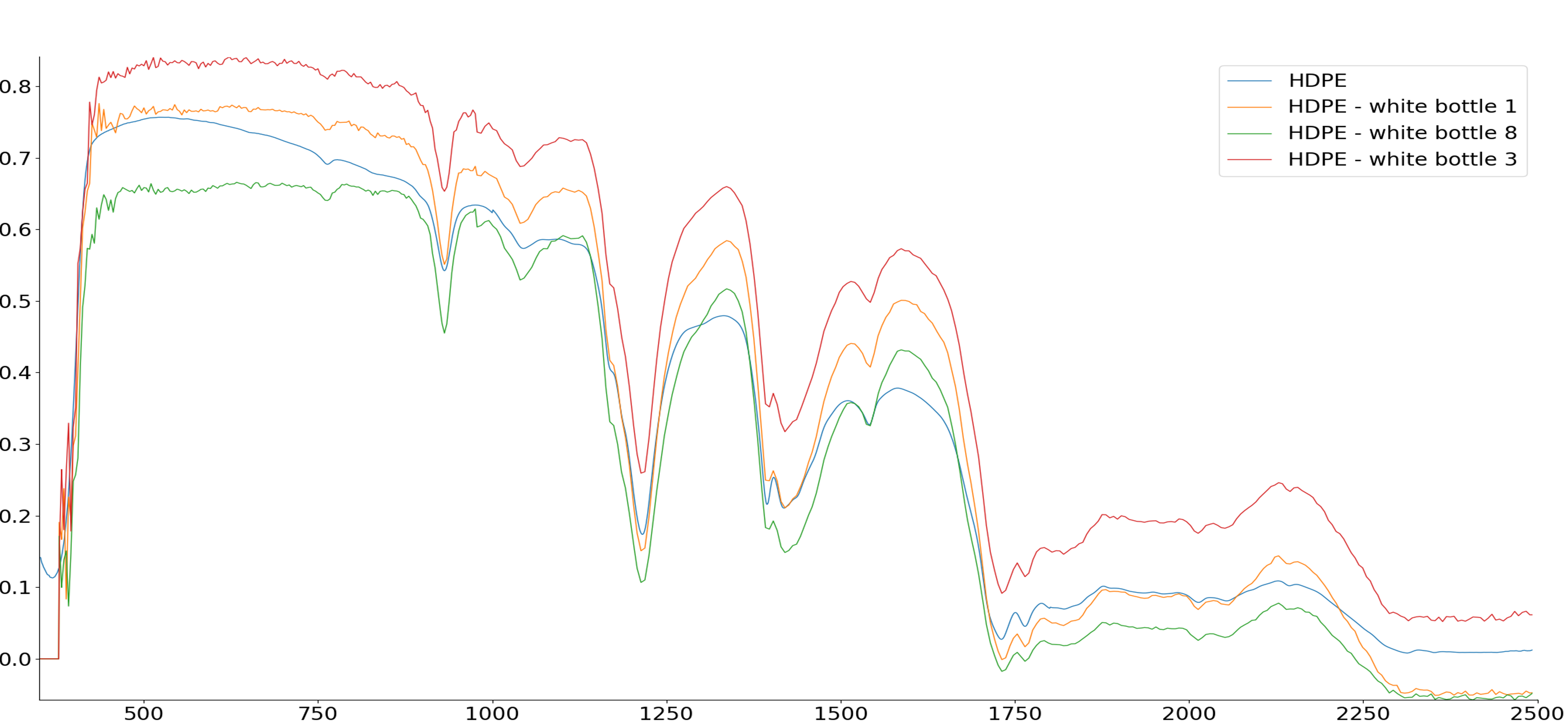}}
\caption{Spectra for five different polymer materials.}
\label{HDPESpectra}
\end{center}
\vskip -0.2in
\end{figure}

The other phenomenology that we care specifically about is that of variation of spectra within a material class.  This variation depends on what is being considered a class; all of the materials whose spectra are shown in~\ref{PolymerSpectra} could be considered part of a 'polymer' class, but they are clearly spectrally distinct.  In figure~\ref{HDPESpectra}, we show multiple spectra from HDPE (high-density polyethylene) and can observe a consistency in the features.

The general material identification problem in hyperspectral imaging can be posed as follows.  We have a pixel from an image and want to know what material(s) is/are present.  We also have a library of spectra of known materials.  The library may have say 1,000 spectra in it.  The observed pixel spectrum is assumed to be a linear mixture of the materials from our library, making this a model selection problem.  We can simplify the problem by assuming that the mixture contains only a few, say four, materials from the library in quantities large enough to be observable, but the problem still has clear challenges.

\section{methods}
A great simplifying component of the identification problem is that we usually do not need the exact set of all materials present in the pixel.  Usually, the question of greatest interest is \textit{is there material from class `X' present in the pixel}.  The class of interest may the class of all polymers; for example searching for lost hikers in a national forest.  The class may be a specific polymer, for example looking for PETE (Polyethylene terephthalate) because this material is commonly used for home greenhouses.  The problem may be less well-defined:  is there a fabric in the pixel, or is there a vehicle present?

Our method for solving this problem is to compute the probability for the class of interest by summing the probabilities for every model that contains a spectrum from that class.  Furthermore, if we label all our spectra into classes, and then select spectra in each class into subclasses, and so on, summing probabilities gives us a probability for every class and subclass, which we call hierarchical probabilistic classification.  Typically, not all spectra are of interest.  If the question is ``Is there a polymer present, and if so what are the probabilities for each different polymer type''  then we just need class and subclass labels for the polymer spectra.  If the question is ``Is there a military-related object in the pixel, then is it a fabric or vehicle, then depending on that get probabilities for further subclassses'' then we need class-subclass-subsubclass labels for all of the military related material spectra.

For a simple example, suppose that we have two spectra, Spectrum 1 and Spectrum 2.  Suppose also that we have a library with 8 spectra: two nylon spectra labeled $N1$ and $N2$, two polyester spectra labeled $P1$ and $P2$, two cotton spectra labeled $C1$ and $C2$, and two vegetation spectra labeled $V1$ and $V2$.  Suppose we compute the probabilities for each spectrum and get the results shown in Table~\ref{Table:ToyExample}.
\begin{table}[htbp]
\caption{Individual probabilities for a toy example.}
\begin{center}
\begin{tabular}{|l|llllllll|}
\hline
      & \multicolumn{6}{c}{Fabric}                                                           & \multicolumn{2}{|c|}{Vegetation} \\
      & \multicolumn{4}{c}{Polymer}                               & \multicolumn{2}{|c}{Cotton} & \multicolumn{2}{|l|}{}           \\
      & \multicolumn{2}{c}{Nylon} & \multicolumn{2}{|c}{Polyester}& \multicolumn{2}{|l}{}       & \multicolumn{2}{|l|}{}           \\
      & N1          & N2          & \multicolumn{1}{|l}{P1}        & P2            & \multicolumn{1}{|l}{C1}           & C2          & \multicolumn{1}{|l}{V1}             & V2            \\  \cline{1-9}
Spec 1 & .4          & .3          & .2            & .1             & 0            & 0           & 0              & 0             \\
Spec 2 & 0           & 0           & 0             & 0              & .1           & 0           & .4             & .5 \\
\hline
\end{tabular}
\label{Table:ToyExample}
\end{center}
\end{table}

In Table~\ref{Table:ToyExample}, we organized the column headers to show the material name for each pair of spectra, and higher level groups such as polymer, which includes the nylon and polyester, and fabrics, which includes the polymer and cotton fabrics.

The purpose of using the probabilities is that for two mutually exclusive events, the probability that one of  the events will occur is the sum of the individual probabilities. Using this, from this table can compute that the probability that Spectrum 1 is a nylon is $0.4+0.3=0.7$.  The probability that Spectrum 1 is polyester is $0.2+0.1=0.3$.  We can compute probabilities for higher level classes: the probability that Spectra 1 is a polymer fabric (eg. nylon or polyester) is $1$, and the probability that Spectrum 1 is a fabric is $1$, and the probability that Spectrum 1 is vegetation is $0$.  These probabilities can be organized in a tree diagram as shown in Figures~\ref{Fig:S1} and~\ref{Fig:S2}.  Assuming that the spectra in the library include all possible materials for these two spectra, we could be confident that Spectrum 1 is a polymer fabric, and more likely a nylon than a polyester.  We could be confident that Spectrum 2 is vegetation.
\begin{figure}
\centering
\begin{minipage}{.4\textwidth}
  \centering
  \Tree [.1\\Library .0.0\\Vegetation [.1.0\\Fabric 0.0\\Cotton [.1.0\\Polymer Polyester\\0.3 Nylon\\0.7 ] !\qsetw{40pt} ] !\qsetw{40pt} ]
  \vspace{0.1in}
  \caption{The material identification tree for Spectrum 1 for the example data.  At each branch in the tree the possible subclasses are listed with probabilities ordered from least likely to most likely.}
  \label{Fig:S1}
\end{minipage}\hspace{.05\textwidth}
\begin{minipage}{.4\textwidth}
\vskip 0.2in
  \centering
  \Tree [.1\\Library 0.1\\Fabric [.0.9\\Vegetation ] !\qsetw{40pt} ]
  \vspace{0.1in}
  \caption{The material identification tree for Spectrum 2 for the example data.  At each branch in the tree the possible subclasses are listed with probabilities ordered from least likely to most likely.}
  \label{Fig:S2}
\end{minipage}
\end{figure}

\section{results}
In this section we provide results from summing probabilities for classes and subclasses.  We first start with a group of pixels from an AVIRIS image and then some targets from the Forest Radiance dataset~\cite{ForestRadiance97}.

\begin{figure}[ht]
\vskip 0.2in
\begin{center}
\centerline{\includegraphics[width=\columnwidth]{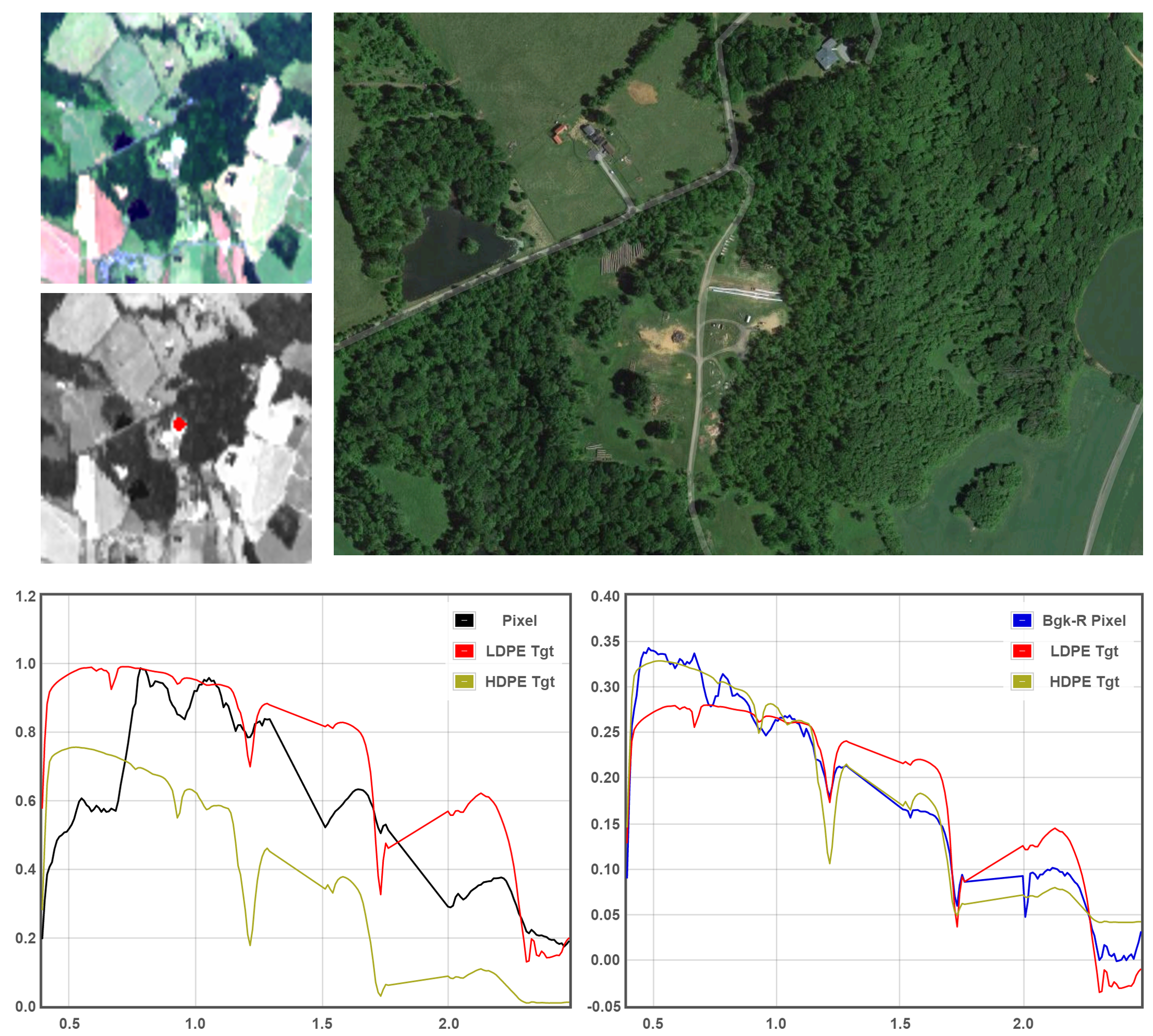}}
\caption{A group so of pixels from the AVIRIS image.}
\label{AVIRISdetect}
\end{center}
\vskip -0.2in
\end{figure}

In Figure~\ref{AVIRISdetect}, we show a detection from an AVIRIS image that was over an area west of Washington DC.  The upper left image is a small `chip' from the image centered on the location, shown as an visual color image by combining the RGB bands from the hyperspectral image.  This location was selected initially selected from the larger image because it scored high with a target detection algorithm~\cite{ManolakisSiracusShaw2001,basener2017geometry}.  Below the RGB chip is a grayscale image with the specific pixels used indicated in red.  (There is a group of pixels that scored high, and we take the average of these spectra as our pixel spectrum for the object.)  The large visual image (right, top) is a high resolution image over this location from Google Earth.  In this image, we can see that there are hay bales wrapped in an LDPE-based material, stacked in lines.  These are likely the source of the detection.  Below these visual images are spectra from this location along with spectra that are known to be from HDPE and LDPE materials for comparison.

In the plots of spectra in Figure~\ref{AVIRISdetect}, the left-hand plot shows the pixel spectrum along with the HDPE and LDPE for comparison.  A standard way to try to confirm materials is to visually examine the spectra and look for features in the pixel that match the know library spectra.  In this plot, it looks like there are some features (small dips or sharp rises) present in this pixel at wavelength locations of the library spectra, but there is not an overall precise match.  This is because the pixel spectrum is a mixture of the material from the wrap as well as other nearby materials, for example the grass, trees, and soil.  In the right-hand plot, these same library HDPE and LDPE spectra are shown, but with these is a plot of the background-removed pixel.  This pixel is computed by selecting a number background spectra from the image near but not on the detected object of interest $s_1,s_2,...,s_n$, computing a linear regression using these background spectra and the spectrum that caused the detection (i.e. the target spectrum),
\begin{equation}
p = a_t s_t + \sum_i a_i s_i
\end{equation}
where $p$ is the pixel spectrum, $s_t$ is the target, and the $a$ are scaler coefficients.  The background-removed spectrum, denoted by Bkg-R, is then
\begin{equation}
\textrm{Bkg-R} = p - \sum_i a_i s_i.
\end{equation}

This background removed spectrum is described in more detail in~\cite{Basener2011}.  We can now see that the background removed pixel spectrum is a good match to both of the polyethylene spectra, but there is a precise match with LDPE among the features in the 2.0-2.5 micrometer range.  (The match depends on the wavelength location where the dips and rises occur, not particularly with overall value.)

The probabilities of inclusion in the model for spectra from a library of about 200 materials, about 50 of which are polymers, are shown in Table~\ref{MaterialProbabilities}.  These were computed from Bayesian model averaging using an Occam's razor search strategy limiting up to 4 elements per model.  It is clear from this table that the LDPE spectra get the highest probabilities, and that the probabilities are spread across these spectra with no one spectrum getting a high probability.  In our terminology, these are overlapping input variables.
\begin{table}[htbp]
\caption{Probabilities for polymers computed from the AVIRIS pixels.}
\begin{center}
\begin{tabular}{|l|l|}
\hline
Material Spectrum   & Probability \\
\hline
LDPE	&	0.21	\\
LDPE - foam packaging 9	&	0.21	\\
LDPE - foam packaging 6	&	0.17	\\
LDPE - foam packaging 3	&	0.12	\\
LDPE - foam packaging 2	&	0.11	\\
LDPE - foam packaging 8	&	0.1	\\
LDPE - foam packaging 5	&	0.09	\\
LDPE - foam packaging 4	&	0.09	\\
LDPE - foam packaging 7	&	0.07	\\
HDPE	&	0.07	\\
HDPE - white bottle 10	&	0.06	\\
HDPE - white bottle 1	&	0.05	\\
HDPE - white bottle 3	&	0.05	\\
HDPE - white bottle 9	&	0.05	\\
HDPE - white bottle 2	&	0.05	\\
PVC - soft inflatable (white) 2	&	0	\\
PVC	&	0	\\
PVC - soft extnesion cord 7	&	0	\\
PVC - soft extnesion cord 3	&	0	\\
PVC - soft extnesion cord 1	&	0	\\
\hline
\end{tabular}
\label{MaterialProbabilities}
\end{center}
\end{table}

\begin{figure}
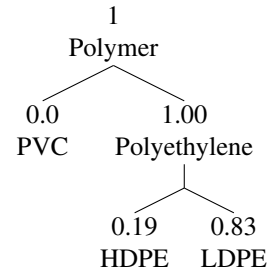

  \Tree [.1\\Polymer 0.0\\PVC [.1.00\\Polyethylene  [0.19\\HDPE 0.83\\LDPE  ] ] ]
  \vspace{0.1in}
  \caption{The class identification tree the pixels detected in the AVIRIS image shown in Figure~\ref{AVIRISdetect}.  At each branch in the tree the possible subclasses are listed with probabilities ordered from least likely to most likely.}
  \label{Fig:AVIRIStree}
\end{figure}

To compute the probabilities for the polymer class, we add the probabilities for all models that included a polymer spectrum, which is 1.0 (there were no models constructed without polymers that passed Occam's window).  To compute probability of PVC vs probability of polyethylene, we added the probabilities of all models that included PVC and all models that included polyethylene, resulting in 0.00 and 1.00 respectively.  Continuing, we arrive with a probability of 0.83 for LDPE and 0.19 for HDPE.  This hierarchal probabilistic identification is depicted as a tree in Figure~\ref{Fig:AVIRIStree}.

\begin{figure}[ht]
\vskip 0.2in
\begin{center}
\centerline{\includegraphics[width=\columnwidth]{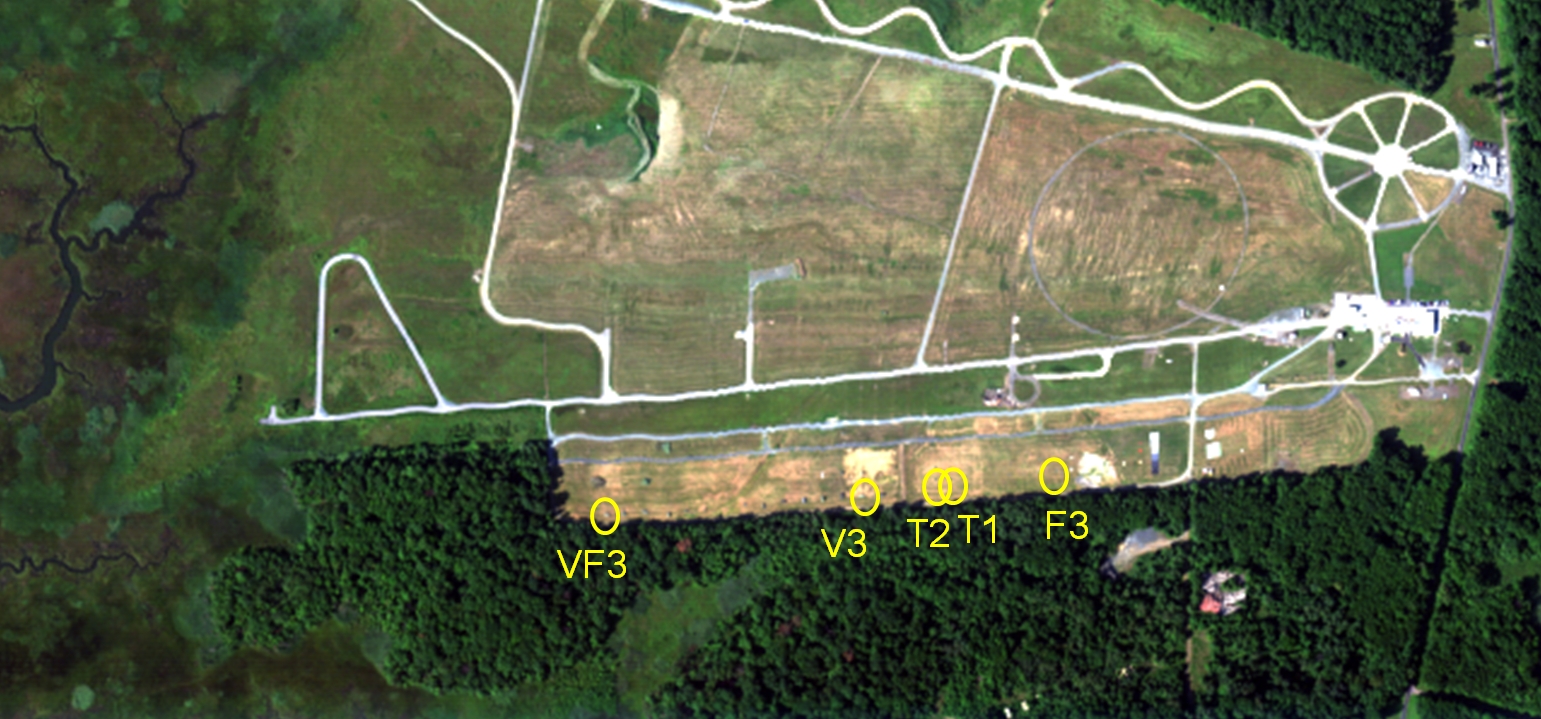}}
\caption{The image forest radiance run09m99 with the selected targets.}
\label{Fig:run09Targets}
\end{center}
\vskip -0.2in
\end{figure}

The next dataset we used is the Forest Radiance image run09m99 acquired in 1995 as part of the HYperspectral Data Imagery Collection Experiment (HYDICE) sensor~\cite{ForestRadiance97,MarianoGrossmann2010}, shown in Figure~\ref{Fig:run09Targets}.  The image was converted to reflectance using the empirical line method computed using the in scene calibration panels.  The image was acquired at approximately 20,000 ft above ground level with a spatial GSD of 3 meters.  Atmospheric and noisy bands were removed leaving 159 bands.  Unlike our AVIRIS example, this image was part of a collection with a number of known target materials placed on the ground with multiple spectra collected for each target.

A number of target vehicles and fabric  panels were placed in the image as part of the experiment.  As part of the experiment 577 spectra were collected off of the targets and background materials.  We will refer to the spectral library consisting of these spectra as the canonical forest radiance library.  The vehicles are designated in 3 classes, V, VF, or DV.  The V class includes vehicles V1 through V7, the VF class includes vehicles VF1 through VF6, and the DV class includes vehicles DV1 through DV4.  Out of the V vehicles, V1-V3 are the same vehicle type and spectra in the library treat this group as a single material class.  The same is true for vehicles VF1-VF3.  The fabric target materials include F1 through F14, T1, T2, and fabrics with other designations. Fabric materials F1-F14, T1, and T2 are emplaced in the image as three square panels each of sizes 1m, 2m, and 3m, arranged in rows.  The fabrics can further be subdivided by material type: F3 is a green cotton material, T1 and T2 and others are green nylon.  The specific objects selected for testing in this paper are the vehicles V3 and VF3 and fabrics F3, T1, and T2, shown in Figure~\ref{Fig:run09Targets}.

At the highest level, one would like to determine if an object is a fabric or a vehicle.  If the determination can be made at that level, we want to determine if a more specific class be designated within vehicle and fabric subclasses.

We first consider the vehicle V3.  The tree diagram shown in Figure~\ref{Fig:V3}.  At the highest level, we assume with probability 1 that the spectra for the selected object is in the library.  This is a basic assumption intrinsic to model averaging.  At the next level, it is determined that the probability that the target is a fabric is 0.0150 while the probability that the target is a vehicle is 0.9850.  Continuing down the tree, we see that there is approximately a 95\% probability that this target is in the class V1-V3, which is the most specific class assignment that can be made with this library.
\begin{figure}
\centering
\begin{minipage}{.4\textwidth}
  \centering
  \Tree [.1\\Library 0.0150\\Fabric [.0.9850\\Vehicle 0.0000\\DV 0.0018\\VF [.0.9834\\V notV1-V3\\0.0420 V1-V3\\0.9580 !\qsetw{40pt} ] !\qsetw{40pt} ] !\qsetw{40pt}  ]
  \vspace{0.1in}
  \caption{The class identification tree for V3.  At each branch in the tree the possible subclasses are listed with probabilities ordered from least likely to most likely.}
  \label{Fig:V3}
\end{minipage}\hspace{.05\textwidth}
\begin{minipage}{.4\textwidth}
\vskip 0.2in
  \centering
  \Tree [.1\\Library 0.1217\\Fabric [.0.8891\\Vehicle 0.0000\\V 0.3175\\DV [.0.5716\\VF notVF1-VF3\\0.1687 VF1-VF3\\0.4029 ] !\qsetw{40pt} ] !\qsetw{40pt} ]
  \vspace{0.1in}
  \caption{The class identification tree for VF3.  At each branch in the tree the possible subclasses are listed with probabilities ordered from least likely to most likely.}
  \label{Fig:VF3}
\end{minipage}
\end{figure}
The class identification tree for vehicle VF3 is shown in Figure~\ref{Fig:VF3}.  Observe that for VF3, it was possible to determine with confidence that this object is a vehicle, and either a DF or VF class vehicle.  After that, the designation becomes less certain.  We can tell that this object is about twice as likely to be a VF than DV, and if it is a VF then it is twice as likely to be a VF1-VF3 than one of the other vehicles.  But this designations within the vehicle classes cannot be made with as much certainty as was possible with V3.

The class identification tree for fabrics T1 and T2 are shown in Figures~\ref{Fig:T1} and~\ref{Fig:T2}.  Observe that for both targets, we can be definitively certain that the fabrics is either T1 or T2.  However, with the T1 target the information is insufficient to determine which of the T-fabrics is correct.  But with the T2 fabric target, we can be confident that the fabric is T2.
\begin{figure}
\centering
\begin{minipage}{.4\textwidth}
  \centering
  \Tree [.1\\Library 0.0000\\Vehicle [.1.0000\\Fabric 0.0000\\Cotton [.1.0000\\Nylon 0.0344\\notT1-2 [.0.96556\\T1-2  0.4759\\T2 0.4898\\T1 ] !\qsetw{40pt} ] !\qsetw{40pt} ] !\qsetw{40pt}  ]
  \vspace{0.1in}
  \caption{The class identification tree for T1.  At each branch in the tree the possible subclasses are listed with probabilities ordered from least likely to most likely.}
  \label{Fig:T1}
\end{minipage}\hspace{.05\textwidth}
\begin{minipage}{.4\textwidth}
\vskip 0.2in
  \centering
  \Tree [.1\\Library 0.0000\\Vehicle [.1.0000\\Fabric 0.0000\\Cotton [.1.0000\\Nylon 0.0268\\notT1-2 [.0.9732\\T1-2  0.0155\\T1 0.9576\\T2 ] !\qsetw{40pt} ] !\qsetw{40pt} ] !\qsetw{40pt}  ]
  \vspace{0.1in}
  \caption{The class identification tree for T2.  At each branch in the tree the possible subclasses are listed with probabilities ordered from least likely to most likely.}
  \label{Fig:T2}
\end{minipage}
\end{figure}

\begin{figure}
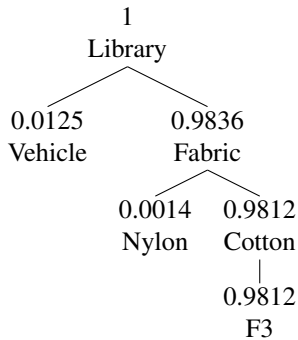

  \Tree [.1\\Library 0.0125\\Vehicle [.0.9836\\Fabric 0.0014\\Nylon [.0.9812\\Cotton [.0.9812\\F3 ]  ]  ] ]
  \vspace{0.1in}
  \caption{The class identification tree for F3, a cotton fabric.  At each branch in the tree the possible subclasses are listed with probabilities ordered from least likely to most likely.}
  \label{Fig:F3}
\end{figure}

The class identification tree for the cotton fabric is shown in Figure~\ref{Fig:F3}.  The final branching only has one branch because F3 is the only known cotton fabric in the library.

\section{conclusions}
In this paper we present a method for computing hierarchical probabilistic classification based on computed probabilities for classes of input regressor variables that represent the same underlying cause or factor in a linear regression problem.  We started by discussing a classical example where police expenditure for two different years is considered in Bayesian model averaging, resulting in the probabilities being distributed over the two years, and leading to manual analysis to understand the general relationship between police expenditure and crime.   We define the term overlapping variables to describe variables that capture the same underlying information, while recognizing that this term is subjective depending on application.

We then provided example computations using hyperspectral imagery.  In this data, the regression problem is determining the linear mixture with abundances of materials present in the pixel of a hyperpsectral image.  However, a given material has multiple different spectra with some variation that might be observed.  Moreover, we might be interested in knowing if any of a class of similar materials is present.  Any two spectra that are measurements for the same material, or for the same class of materials, would be considered overlapping variables.

In our example from the AVIRIS hyperspectral image, we compute the probability that a polymer of any type is present, then the probability between PVC and polyethylene, and then final between high density polyethylene (HDPE) vs low density polyethylene (LDPE).  The computation indicates that the material is a polyethylene with probability 1, and more likely LDPE (p=0.83) rather than HDPE (p=0.19).  We then produce examples using imagery from the HYDICE sensor in the Forest Radiance dataset.

This method provides certainly more information than a pick-a-winner method for model selection that simply chooses a single best model.  Pick-a-winner approaches ignore the uncertainty present with multiple good models.  Bayesian model averaging estimates the uncertainty among models in the probabilities for individual regressor variable, but as we observed that overlapping variables can create a situation when the individual variable probabilities can underestimate the probability for a single underlying factor/cause that is represented by multiple different variables.  Our method computes probabilities for underlying factors/causes, and can do so simultaneously at multiple levels of class resolution (classes, subclasses, etc.).  We display the results in trees with class labels and probabilities at each branch point.
\bibliography{my_refs3}
\bibliographystyle{IEEEbib}

\end{document}